\documentclass[letterpaper, 10 pt, journal, twoside]{ieeetran}
\usepackage[utf8]{inputenc} 
\usepackage[T1]{fontenc}    
\usepackage{hyperref}       
\usepackage{url}            
\usepackage{booktabs}       
\usepackage{amsfonts}       
\usepackage{nicefrac}       
\usepackage{microtype}      
\usepackage{xcolor}         

\usepackage{bm}
\usepackage{amsmath}
\usepackage{amssymb}
\usepackage{mathtools}
\usepackage{amsthm}
\usepackage{caption}
\usepackage{multirow}
\usepackage{makecell}
\usepackage{fontawesome}
\usepackage{marvosym}
\usepackage[top=57pt, bottom=43pt, left=48pt,right=48pt]{geometry}
\usepackage{graphicx}
\usepackage{wrapfig}
\usepackage{subcaption}

\usepackage[ruled,vlined]{algorithm2e}
\usepackage{algorithmicx}
\usepackage{algpseudocode}
\usepackage{makecell}
\newcommand{\name}{{\textrm{OPEN}}}  

\begin{document}

\title{Online Planning for Multi-UAV Pursuit-Evasion in Unknown Environments Using Deep Reinforcement Learning}

\author{Jiayu Chen$^{1*}$, 
Chao Yu$^{1*\textsuperscript{\Letter}}$,
Guosheng Li$^{1}$,
Wenhao Tang$^{2}$,
Shilong Ji$^{1}$,
Xinyi Yang$^{1}$,\\
Botian Xu$^{1}$,
Huazhong Yang$^{1}$,
Yu Wang$^{1}$\textsuperscript{\Letter}
\thanks{Manuscript received: March, 4, 2025; Revised May, 17, 2025; Accepted June, 16, 2025.}
\thanks{This paper was recommended for publication by Editor M. Ani Hsieh upon evaluation of the Associate Editor and Reviewers' comments.}
\thanks{This research was supported by National Natural Science Foundation of China (No.62406159, 62325405), Postdoctoral Fellowship Program of CPSF under Grant Number (GZC20240830, 2024M761676), China Postdoctoral Science Special Foundation 2024T170496.}
\thanks{* Equal Contribution.~\url{jia768167535@gmail.com}~{\Letter} Corresponding Authors. \url{zoeyuchao@gmail.com} and \url{yu-wang@tsinghua.edu.cn}}
\thanks{$^{1}$Tsinghua University, Beijing, 100084, China. $^{2}$Tsinghua Shenzhen International Graduate School,
Shenzhen, 518055, China.}
\thanks{Digital Object Identifier (DOI): see top of this page.}}

\maketitle

\begin{abstract}
Multi-UAV pursuit-evasion, where pursuers aim to capture evaders, poses a key challenge for UAV swarm intelligence. Multi-agent reinforcement learning (MARL) has demonstrated potential in modeling cooperative behaviors, but most RL-based approaches remain constrained to simplified simulations with limited dynamics or fixed scenarios. Previous attempts to deploy RL policy to real-world pursuit-evasion are largely restricted to two-dimensional scenarios, such as ground vehicles or UAVs at fixed altitudes. In this paper, we propose a novel MARL-based algorithm that learns \underline{o}nline planning for multi-UAV \underline{p}ursuit-\underline{e}vasion in u\underline{n}known environments ({\name}). {\name} introduces an evader prediction-enhanced network to tackle partial observability in cooperative policy learning. Additionally, {\name} proposes an adaptive environment generator within MARL training, enabling higher exploration efficiency and better policy generalization across diverse scenarios. Simulations show our method significantly outperforms all baselines in challenging scenarios, generalizing to unseen scenarios with a 100\% capture rate. Finally, after integrating calibrated dynamics models of UAVs into training, we derive a feasible policy via a two-stage reward refinement and deploy the policy on real quadrotors in a zero-shot manner. To our knowledge, this is the first work to derive and deploy an RL-based policy using collective thrust and body rates control commands for multi-UAV pursuit-evasion in unknown environments. The open-source code and videos are available at \url{https://sites.google.com/view/pursuit-evasion-rl}.

\end{abstract}

\section{INTRODUCTION}\label{sec:intro}
\IEEEPARstart{M}{ulti}-UAV pursuit-evasion, where teams of pursuers attempt to capture evaders while the evaders employ evasive strategies to avoid capture, is a key application of UAV swarms. Multi-UAV pursuit-evasion has important applications in both military and civilian contexts, including UAV defense systems~\cite{turetsky2003missile}, adversarial drone engagements~\cite{eklund2011switched}, and search-and-rescue~\cite{oyler2016pursuit}.

Traditional approaches to solving pursuit-evasion games, such as game theory~\cite{vidal2002probabilistic}, control theory~\cite{ye2020satellite,fang2020cooperative,tian2021distributed,huang2011guaranteed,pierson2016intercepting,palm2013particle}, and heuristic methods~\cite{angelani2012collective,janosov2017group,koren1991potential}, face significant limitations when applied to complex real-world scenarios. These methods require accurate and reasonable knowledge of models and initial conditions~\cite{mu2023survey}, which may struggle to handle the nonlinear dynamics and high-dimensional environments typically encountered in practical applications. As a result, researchers have increasingly turned to artificial intelligence (AI) techniques, with reinforcement learning (RL) emerging as a leading solution\cite{gupta2017cooperative,desouky2011q,awheda2015residual,luo2024multi,gan2023multi}. RL enables UAVs to iteratively learn pursuit and evasion strategies by interacting with simulated environments. By leveraging neural networks, RL can model complex, cooperative behaviors and discover strategies that are difficult to encode using explicit rules.

However, despite these advancements, most RL-based methods for pursuit-evasion focus on simplified tasks in simulation~\cite{wang2020cooperative,xu2020multi,huttenrauch2019deep,kokolakis2022safety,liu2024game}. These methods frequently model agents, both pursuers and evaders, as point masses or with limited kinematic properties, developing only high-level strategies. Such abstractions overlook the kinematic and dynamic constraints of real-world systems, limiting the effectiveness of these strategies outside simulation environments. Moreover, many RL approaches are tailored to fixed, predefined scenarios, reducing their ability to generalize to diverse and unknown scenarios~\cite{gupta2017cooperative,wang2020cooperative,zhang2022multi}. Although recent work has explored deploying RL-based pursuit-evasion strategies in real-world settings~\cite{zhang2022game,zhang2022multi,pierson2016intercepting,de2021decentralized}, these efforts have primarily been restricted to two-dimensional tasks, such as ground vehicles or UAVs constrained to fixed altitudes, without addressing the complexities of real three-dimensional environments.

This paper aims to learn an RL policy for multi-UAV pursuit-evasion, perform online planning in unknown environments, and deploy it on real UAVs. The problems are:
\begin{itemize}
    \item \textbf{Joint optimization of planning and control}: UAV actions must be coordinated to capture the evader under partial observation, while avoiding environmental obstacles, preventing collisions, and adhering to dynamics
    model and physical constraints for safe and feasible flight.
    \item \textbf{Large exploration space}: The 3D nature of UAVs, combined with varying scenarios, significantly expands the state space, resulting in a large number of samples required to find viable strategies using RL.
    \item \textbf{Policy generalization}: RL strategies that are optimized for specific scenarios often fail to generalize to new environments. 
    \item \textbf{Sim-to-real transfer}: The sim-to-real gap, a common issue in RL, is particularly pronounced in multi-UAV pursuit-evasion tasks due to the physical constraints of UAVs and the need for agile, precise maneuvers. 
\end{itemize}

We address these challenges by integrating calibrated dynamics models of UAVs into training policy for multi-UAV pursuit-evasion tasks. The RL-based policy generates collective thrust and body rates (CTBR) control commands, balancing flexibility, cooperative decision-making, and sim-to-real transfer. The overall algorithm is named {\name} (\underline{O}nline planning for multi-UAV \underline{p}ursuit-\underline{e}vasion in u\underline{n}known environments). {\name} proposes an attention-based, evader-prediction-enhanced network that integrates predictive information about the evader's movements into the RL policy inputs, improving the ability to cooperatively capture the evader with partial observation. Additionally, we introduce an adaptive environment generator to MARL training, which automatically generates diverse and appropriately challenging curricula. This boosts sample efficiency and enhances policy generalization to unseen scenarios. Finally, we employ a two-stage reward refinement process to regularize policy output and ensure stability of behavior during real-world deployment.

We evaluate performance in four test scenarios, with results showing that {\name} outperforms all baselines with a clear margin and maintains a 100\% capture rate in unseen scenarios, demonstrating strong generalization. Ablation studies show our approach improves sample efficiency by over 50\% in the training tasks. Our method also demonstrates robustness to the evader speed, maintaining a high capture success rate even when confronted with high-speed evaders that are never encountered during training. We also deploy our policy on real Crazyflie quadrotors in a zero-shot manner, achieving strategies in the real world that closely mirror those observed in simulation. To the best of our knowledge, this is the first RL policy with CTBR commands in pursuit-evasion tasks and can be deployed directly on real quadrotors in unknown environments.
Our contributions can be summarized as follows:
\begin{itemize}
\item We propose {\name}, a novel MARL-based algorithm for online planning in multi-UAV pursuit-evasion tasks in unknown environments. It uses collective thrust and body rates control commands as policy outputs, ensuring flexibility and enabling zero-shot transfer to real-world.

\item We introduce an attention-based evader-prediction network that integrates predicted evader trajectories into policy inputs to guide cooperative strategies under partial observations, alongside an adaptive environment generator that generates challenging curricula to boost sample efficiency and generalization to unseen scenarios. We integrate a calibrated UAV dynamics model into training and employ a two-stage reward refinement process to ensure RL policy stability for real-world deployment.

\item Simulation experiments demonstrate that our method achieves near 100\% capture rates across four test scenarios, outperforming all baseline approaches. Furthermore, we successfully deploy the multi-UAV pursuit policy on real Crazyflie quadrotors in a zero-shot manner.
\end{itemize}

\section{RELATED WORK}\label{sec:related}
In pursuit-evasion games, one or more pursuers aim to capture one or more evaders, while the evaders actively avoid capture, distinguishing this from search strategies where the target is passive~\cite{isler2005randomized}. Traditional research in this area can be broadly categorized into game-theoretic, control-theoretic, and heuristic approaches~\cite{mu2023survey}. Game-theoretic methods simplify the problem into mathematical models, primarily focusing on differential games~\cite{vidal2002probabilistic}. Control-theoretic approaches model the nonlinear dynamics of the pursuit, optimizing strategies using the Hamilton-Jacobi-Isaacs equation~\cite{ye2020satellite,fang2020cooperative,tian2021distributed}. Heuristic methods design forces to guide pursuers—such as attraction to the evader and repulsion from teammates and obstacles—and often utilize optimization techniques like particle swarm optimization (PSO)~\cite{palm2013particle} and artificial potential fields (APF)~\cite{koren1991potential}. However, these traditional methods have limitations in real-world applications due to the gap between their simple system modeling and real UAVs, and the difficulty of handling complex cooperative strategies.

Reinforcement learning (RL) has emerged as a powerful data-driven approach for solving multi-agent coordination problems in pursuit-evasion games~\cite{zhang2022multi,wang2020cooperative,xu2020multi,huttenrauch2019deep,kokolakis2022safety,liu2024game, cheng2024multi}. Several studies have applied RL to enhance traditional methods~\cite{zhang2022multi} or address issues such as varying agent numbers~\cite{xu2020multi}, efficient communication~\cite{wang2020cooperative}, credit assignment~\cite{10269079} or perception uncertainty~\cite{cheng2024multi}. However, most RL research remains limited to simplified simulation environments, where agents are modeled as particles or constrained by limited kinematics, reducing the effectiveness of these strategies in real-world scenarios. While recent work has explored deploying RL-based strategies in pursuit-evasion tasks~\cite{zhang2022game,zhang2022multi,pierson2016intercepting,de2021decentralized}, these efforts have largely been confined to two-dimensional environments, such as ground vehicles, surface vehicles or UAVs at fixed altitudes.

In this paper, we consider the UAV dynamics model and physical constraints, introducing an evader-prediction enhanced network and an adaptive environment generator to address the challenges of large exploration space and policy generalization to unseen scenarios. The learned RL policy is capable of zero-shot deployment to real quadrotors.

\section{PRELIMINARY}\label{sec:prelim}
\begin{figure}[t]
\begin{minipage}{0.48\textwidth}
\centering
\subcaptionbox{Partially observable scenario\label{fig:schema}}{\includegraphics[width=0.65\textwidth]{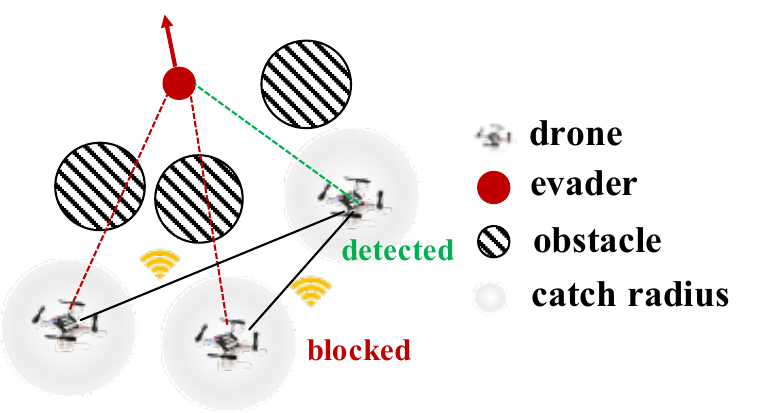}}
\subcaptionbox{Sim. environment\label{fig:snapshot}}{\includegraphics[width=0.32\textwidth]{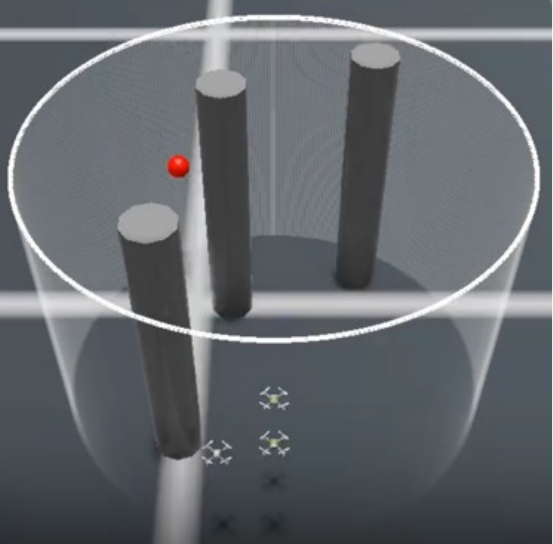}}
\caption{Setup of multi-UAV pursuit-evasion.}
\vspace{-20pt}
\end{minipage}
\end{figure}

\begin{figure*}[t]
\centering
\includegraphics[width=0.7\textwidth]{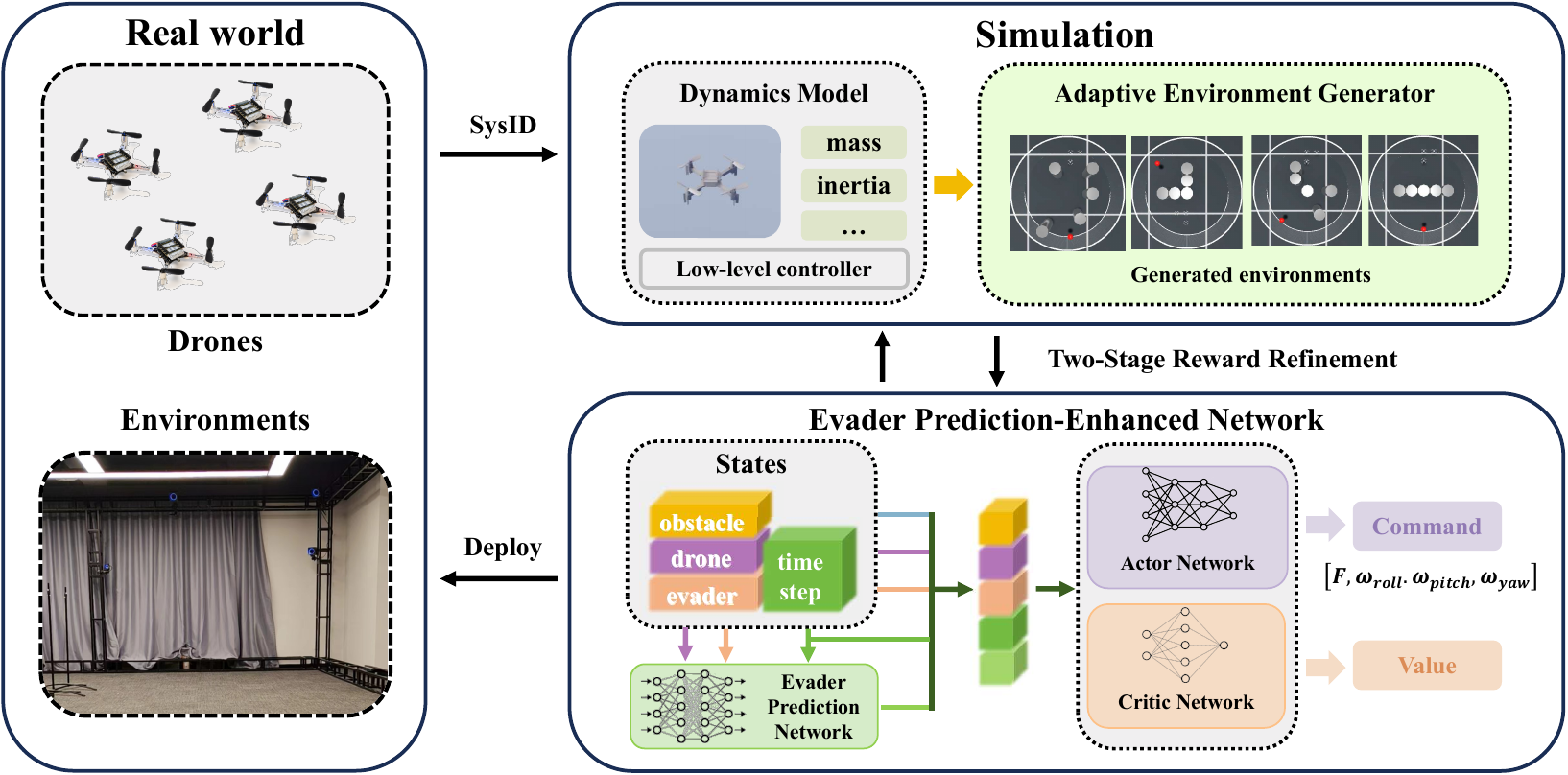}
\caption{Pipeline of our method. We begin by calibrating the parameters of the quadrotor dynamics model through system identification. Next, we introduce an \emph{Adaptive Environment Generator} to automatically generate tasks for policy training and employ an \emph{Evader Prediction-Enhanced Network} for cooperative capture under partial observations. Finally, with two-stage reward refinement, the learned policy is directly transferred to real quadrotors in a zero-shot manner.}
\vspace{-10pt}
\label{fig:overview}
\end{figure*}

\subsection{Pursuit-evasion Problem}
As shown in Fig.~\ref{fig:schema}, the multi-UAV pursuit-evasion problem involves $N$ UAVs chasing a faster evader in an obstacle-filled environment. The UAVs aim to capture the evader quickly while avoiding obstacles. The evader is captured if it comes within the capture radius of a UAV. Detection is possible only if no obstacles block the line of sight. When detected, the UAV can pass information to its teammates. The non-strategic evader, controlled using a potential field method \cite{de2021decentralized}, experiences repulsive forces from UAVs, obstacles, and arena boundaries, with the magnitude of the force inversely proportional to the distance. The evader moves at a constant speed of $v_e$, with its direction determined by the resultant of these repulsive forces.
\vspace{-3pt}

\subsection{Problem Formulation}
We formulate the multi-UAV pursuit-evasion task as decentralized, partially observable Markov
decision processes (Dec-POMDPs), $M=<\mathcal{N},\mathcal{S},\mathcal{A},\mathcal{O}, \mathcal{S}_0, \mathcal{P}, \mathcal{R},\gamma>$ with the state space $\mathcal{S}$, the joint action space $\mathcal{A}$, the observation space $\mathcal{O}$, the space of initial states $\mathcal{S}_0$, the transition probability $\mathcal{P}$, reward function $\mathcal{R}$ and the discount factor $\gamma$. $\mathcal{N}\equiv\{1,...,N\}$ is a set of $N$ UAVs. 
$o_i=\mathcal{O}(s;i)$ denotes the observation for the UAV $i$ under state $s\in\mathcal{S}$. 

The goal of our work is to construct a policy that performs well across diverse scenarios. The task space $\mathcal{T}$ defines a series of Dec-POMDPs with similar properties. We use a parameter space $\mathcal{W}$ to represent the inter-task variation of $\mathcal{T}$. We generate the corresponding initial states by a task parameter $w\in \mathcal{W}$. In our setup, $w$ comprises the initial positions of the UAVs and the evader, as well as the number and positions of obstacles. 
We consider homogeneous UAVs and learn a parameter-sharing policy $\pi_\theta$ parameterized by $\theta$ to output action $a_i\sim \pi_\theta( \cdot
|o_i)$ for UAV $i$. The final objective $J(\theta)$ is to maximize the expected accumulative reward for any task parameter $w\in\mathcal{W}$, i.e.,
$J(\theta)=\mathbb{E}_{w\sim\mathcal{W}}\left[\sum_t \gamma^t R(s^t,\boldsymbol{a}^t; w) \mid w\right]$, where $\boldsymbol{a}^t$ is joint actions at timestep $t$.

\subsection{Quadrotor Model}
In this paper, we use quadrotors, one of the most widely used UAV, as our physical platform. The quadrotor is assumed to be a $6$ degree-of-freedom rigid body of mass $m$ and diagonal moment of inertia matrix $\textbf{\textit{I}} = diag(I_x, I_y, I_z)$. The dynamics of the quadrotor are modeled by the differential equation: $\boldsymbol{\dot{x}} = [\boldsymbol{\dot{p}}^T, \boldsymbol{\dot{q}}^T, \boldsymbol{\dot{v}}^T, \boldsymbol{\dot{\omega}}^T],$ where the quadrotor state $\boldsymbol{x}\in\mathbb{R}^{13}$ consists the position $\boldsymbol{p}$, the orientation $\boldsymbol{q}$ in quaternions, the linear velocity $\boldsymbol{v}$ and the angular velocity $\boldsymbol{\omega}$. The acceleration of the quadrotor is described as 
\begin{equation}
    \boldsymbol{\dot{v}} = \begin{bmatrix}
0 \\
0 \\
-g
\end{bmatrix} + R\begin{bmatrix}
0 \\
0 \\
\sum_{j} \boldsymbol{f}_j / m
\end{bmatrix},
\end{equation}
in the world frame with gravity $g$. $\boldsymbol{f}_j$ is the force generated by the $j$-th rotor. $R$ is the rotation matrix from the body frame to the world frame. 

The angular acceleration calculated by Euler's rotation equation in the body frame is $\boldsymbol{\dot{\omega}} = \boldsymbol{I}^{-1}(\boldsymbol{\tau} - \boldsymbol{\omega}\times(\boldsymbol{I}\boldsymbol{\omega})).$ The torques $\boldsymbol{\tau}$ acting in the body frame are determined by $\boldsymbol{\tau} = \sum_{j}\boldsymbol{\tau}_j + \boldsymbol{r}_{pos, j}\times \boldsymbol{f}_j,$
where $\boldsymbol{\tau}_j$ is the torques generated by the $j$-th rotor, $\boldsymbol{r}_{pos, j}$ is the position of the rotor $j$ expressed in the body frame. We model the rotational speeds of the four motors $\Omega_j$ as first-order system with time constant $T_m$ where the commanded motor speeds $\Omega_{cmd}$ are the input, i.e., $\dot{\Omega_j} = T_m(\Omega_{cmd,j} - \Omega_j).$ The force and torque produced by the $j$-th rotor are modeled as follows: $\boldsymbol{f}_j = k_f{\Omega_j}^2, \boldsymbol{\tau}_j = k_m{\Omega_j}^2.$

\section{METHODOLOGY}\label{sec:method} 
Our method's pipeline is illustrated in Fig.~\ref{fig:overview}. To bridge the sim-to-real gap, we calibrate quadrotor dynamics via system identification and integrate them into a GPU-parallel simulator \cite{xu2023omnidrones}. We use collective thrust and body rates (CTBR) as control commands and train RL policy to output it via a SOTA MARL algorithm, MAPPO~\cite{yu2022mappo}. To model the evader's future trajectory and guide cooperative strategies under partial observability, we design an \emph{Evader Prediction-Enhanced Network} that leverages an attention-based architecture to capture the interrelations within observations and a trajectory prediction network to forecast the evader's movement. To further enhance policy generalization to different scenarios, we propose an \emph{Adaptive Environment Generator} to automatically generate appropriately challenging curricula, enabling efficient exploration of the entire task space. Finally, with the two-stage reward refinement, the learned policy is applied directly to real quadrotors without any real data fine-tuning. The following sections detail the MARL setup, the \emph{Evader Prediction-Enhanced Network}, the \emph{Adaptive Environment Generator}, and the two-stage reward refinement.

\subsection{Multi-Agent Reinforcement Learning Setup}
\textbf{Observation Space.} The observation $\mathbf{o}_i$ for quadrotor $i$ is composed of three components: the self-information $\mathbf{o}_{self}$, states of other quadrotors $\mathbf{o}_{other}$ and information about obstacles $\mathbf{o}_{ob}$.
$\mathbf{o}_{self}$ consists of its orientation in quaternions, the linear velocity, the real and predicted future relative position to the evader. 
If the evader is not detected by any of the quadrotors, we mask the real relative position with a fixed marker value which is $-5$ in our setup. $\mathbf{o}_{other}$ contains the relative positions to other quadrotors. $\mathbf{o}_{ob}$ denotes the relative positions to the $k$-nearest obstacles, where $k$ is smaller than the maximum number of obstacles, representing the quadrotor's inability to sense global information. In our setup, $k=3$.

\textbf{Action Space.} To achieve agile control and facilitate highly cooperative decision-making, we employ collective thrust and body rates (CTBR) commands as the policy actions. These CTBR commands are subsequently translated into precise motor inputs by a low-level PID controller. Specifically, the action for quadrotor $i$ is defined as $a_i = (F, \omega_{\text{roll}}, \omega_{\text{pitch}}, \omega_{\text{yaw}})$, where $F \in [0, 1]$ represents the normalized collective thrust, and $\omega \in [-\pi, \pi]~rad/s $ denotes the body rates along the roll, pitch, and yaw axes, respectively. 

\textbf{Reward Function.} 
The team-based reward function promotes cooperative evader capture and obstacle avoidance, comprising four components: capture reward, distance reward, collision penalty, and smoothness reward. The capture reward incentivizes teamwork, granting all quadrotors a +6 bonus if the evader enters any quadrotor's capture radius. The distance reward, with a coefficient of $-0.1$, penalizes quadrotors based on their distance to the evader, encouraging closer proximity. A collision penalty of $-10$ is imposed if quadrotors breach a safety threshold near obstacles, ensuring safe capture. The smoothness reward regularizes policy outputs to ensure reliable behavior during deployment, as detailed in Sec.~\ref{sec:reward-refine}.

\subsection{Evader Prediction-Enhanced Network}
In the evader prediction-enhanced network, we build an \emph{Evader Prediction Network} that predicts the evader's future trajectories based on historical observations, facilitating the learning of cooperative pursuit strategies under partial observation. The predicted trajectories, concatenated with the raw observations, are then fed into the \emph{Actor and Critic Networks}. These networks utilize an attention-based observation encoder to better capture the relationships between different entities.

\begin{figure}[t]
\centering
\includegraphics[width=0.45\textwidth]{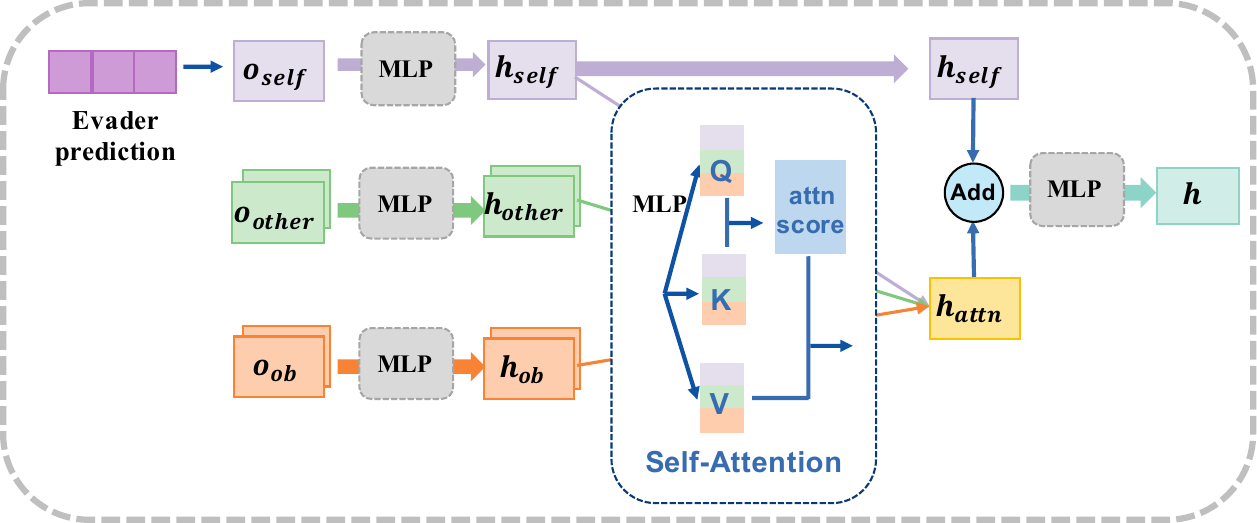}
\caption{Attention-based observation encoder.}\label{fig:nn}
\end{figure}
\subsubsection{Evader Prediction Network}
We use an LSTM to predict the evader's trajectory for the next $K$ timesteps. The input consists of $n$-step historical data, including the positions of all quadrotors, the positions and velocities of the evader, and the current timestep. If the evader is blocked by an obstacle and undetected by any quadrotor, we use a marker value to replace the evader's positions and velocities. To train the network, we collect data over $n+K$ timesteps during the rollout phase, using the first $n$ timesteps as inputs $X$ and the subsequent $n+1$ to $n+K$ timesteps as labels $Y$. The evader prediction network $P_{\phi}$ is trained via supervised learning, and the loss is defined as $L_{\phi} = \mathbb{E}(Y - P_{\phi}(X))^2$.

\subsubsection{Actor and Critic Networks}
The actor and critic networks are based on the attention-based observation encoder, as illustrated in Fig.~\ref{fig:nn}. The raw observation of quadrotor $ i $ consists of three components: $\mathbf{o}_{self}$, $\mathbf{o}_{other}$, and $\mathbf{o}_{ob}$. The predicted trajectory of the evader is concatenated into $\mathbf{o}_{self}$. Each component is separately encoded using distinct MLPs, producing embeddings of $128$ dimensions each. A multi-head self-attention module is employed to capture the relationships between these embeddings, resulting in features $\mathbf{h}_{attn}$. To emphasize self-information, the self-embeddings $\mathbf{h}_{self}$ are added to $\mathbf{h}_{attn}$ and passed through an MLP to generate the final feature $\mathbf{h}$. In the actor network, actions are parameterized using a Gaussian distribution based on $\mathbf{h}$. For the critic network, $\mathbf{h}$ is input into an MLP to produce a scalar value representing the estimated state value.
\begin{algorithm}[t]
 \caption{Training with Adaptive Environment Generator}
 \label{algo:full}
   \textbf{Input:} {$\theta$, $p$, $\mathcal{Q}_{active}$, $\sigma_{min}$, $\sigma_{max}$}\; 
   
  \textbf{Output:} final policy $\pi_{\theta}$\;
  $\mathcal{Q}_{active}\gets\{\}$\;
  \Repeat{$\pi_{\theta}$ converges}
  {
   \tcp{Local expansion and global exploration.}
    $W_{env}\gets$ sample tasks from $\mathcal{Q}_{active}$ with probability $p$ and obtain expanded tasks $w$ by $\textbf{Expand}$, else $w \sim \mathrm{Unif}(\mathcal{W})$ with probability $1-p$\;
    \tcp{Train the policy $\pi_{\theta}$.}
    Train and evaluate $\pi_{\theta}$ with $W_{env}$ via MARL\;
    \tcp{select task parameters from $W_{env}$}
    $W_{new}\gets\textbf{Selection}(W_{env}, \sigma_{min}, \sigma_{max})$\;
    
    Add $W_{new}$ to the active archive $\mathcal{Q}_{active}$\;
  }
 \end{algorithm}

 \begin{figure}
 \centering
\includegraphics[width=0.4\textwidth]{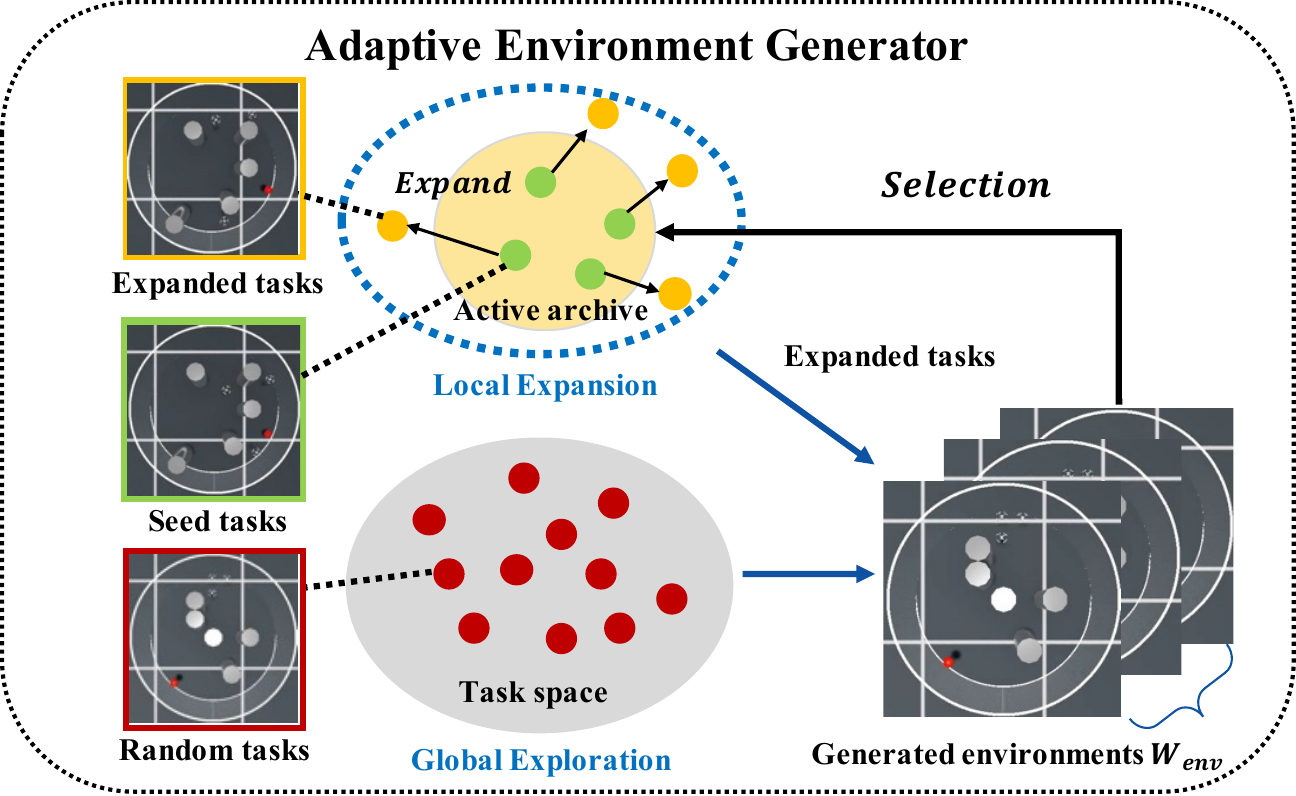}
\caption{Design of the \emph{Adaptive Environment Generator}.}
\label{fig:envgen}
\vspace{-5mm}
\end{figure}
 
\subsection{Adaptive Environment Generator}
Our goal is to derive an RL policy that generalizes effectively to unknown environments. While domain randomization \cite{tobin2017domain} is a common technique for improving generalization, its reliance on randomizing task parameters often requires a large number of samples to achieve reliable results. This approach becomes sample-inefficient and struggles with complex or rare corner cases, where underrepresented environmental configurations lead to suboptimal performance in unseen scenarios (see Sec. \ref{sec:ab}). In contrast, our adaptive environment generator efficiently explores the task space by focusing on corner cases and rare scenarios, integrating this dynamic approach into MARL training, as detailed in Algo. \ref{algo:full}.

We leverage two inductive biases: (1) different obstacle configurations require distinct pursuit strategies, and (2) under the same obstacle setup, the initial positions of UAVs and the evader significantly affect capture difficulty. Based on these insights, we decompose the task space into two modules: \emph{Local Expansion}, which explores UAV and evader positions under fixed obstacles, and \emph{Global Exploration}, which investigates diverse obstacle configurations to generate simulation environments (Fig.~\ref{fig:envgen}).

The \emph{Local Expansion} module enhances the quadrotors' ability to capture the evader from any initial position within a fixed obstacle setup. It maintains an active archive $\mathcal{Q}_{active}$ of task parameters $w$ for unsolved environments. Tasks are expanded from seed tasks sampled from $\mathcal{Q}_{active}$ by adding noise $\epsilon\in[-\delta,\delta]$ to UAV and evader positions while keeping obstacles unchanged. The archive is updated by evaluating the policy in generated environments $W_{env}$ and retaining tasks with success rates $c \in [\sigma_{min}, \sigma_{max}]$, where $\sigma_{min} = 0.5$ and $\sigma_{max} = 0.9$. This process gradually increases task complexity, enabling automatic curriculum generation.

The \emph{Global Exploration} module explores unseen obstacle configurations by randomly sampling task parameters from the entire parameter space $\mathcal{W}$, including the number and positions of obstacles, as well as the initial positions of the UAVs and the evader. To balance focused and diverse task generation, we combine fast \emph{Local Expansion} (selected with a probability of $p=0.7$) and slow \emph{Global Exploration} (selected with a probability of $1-p=0.3$).

\subsection{Two-stage Reward Refinement}\label{sec:reward-refine}
We use the smoothness reward $e^{-||a^t - a^{t-1}||_2}$, where $\boldsymbol{a}_t$ represents the policy's CTBR output, to balance task completion and action smoothness~\cite{chen2024matters}. To address the challenge of maintaining a feasible policy for real-world deployment while ensuring task success, we employ a two-stage reward refinement process. In the first stage, we exclude the smoothness reward and focus on the other three rewards. In the second stage, we reintroduce the smoothness reward and continue training from the first-stage checkpoint. This ensures high task success while improving action smoothness, with the smoothness reward coefficient set to $2.0$.

\section{EXPERIMENT}\label{sec:expr}
\begin{table*}[t]
\centering
{
{\begin{tabular}{ccccc|cc|c}
\toprule
Scenarios& Metrics&Angelani&Janosov &APF  & DACOOP &MAPPO & {\name}  \\ 
\midrule
\multirow{4}{*}{\emph{Wall}} & \textit{Cap. Rate}$\uparrow$ & 0.008\scriptsize{(0.003)} & 0.009\scriptsize{(0.002)} & 0.010\scriptsize{(0.005)} & 0.205\scriptsize{(0.007)} &  0.817\scriptsize{(0.095)} & \textbf{0.987\scriptsize{(0.019)}}     \\ 
\cmidrule{2-8}
& \textit{Cap. Step}$\downarrow$  & 798.0\scriptsize{(000.8)} & 797.0\scriptsize{(000.8)} & 797.0\scriptsize{(001.4)}   & 744.0\scriptsize{(009.1)}& 569.1\scriptsize{(046.9)} & \textbf{480.4\scriptsize{(052.4)}}    \\ 
\cmidrule{2-8}
& \textit{Coll. Rate}$\downarrow$  & 0.010\scriptsize{(0.000)} & 0.010\scriptsize{(0.000)} & 0.010\scriptsize{(0.000)}   & 0.040\scriptsize{(0.008)}& 0.010\scriptsize{(0.076)} & \textbf{0.000\scriptsize{(0.000)}}    \\ 
\cmidrule{1-8}
\multirow{4}{*}{\emph{Narrow Gap}} & \textit{Cap. Rate}$\uparrow$ & 0.213\scriptsize{(0.012)} &0.219\scriptsize{(0.012)} & 0.253\scriptsize{(0.012)} &0.447\scriptsize{(0.017)} &  0.917\scriptsize{(0.034)} & \textbf{1.000\scriptsize{(0.000)}}     \\ 
\cmidrule{2-8}
& \textit{Cap. Step}$\downarrow$   & 755.0\scriptsize{(004.5)} & 757.7\scriptsize{(004.2)} &  743.0\scriptsize{(002.2)}   & 553.0\scriptsize{(002.2)}& \textbf{484.0\scriptsize{(041.2)}} & 526.0\scriptsize{(032.9)}    \\ 
\cmidrule{2-8}
& \textit{Coll. Rate}$\downarrow$   & 0.094\scriptsize{(0.001)} & 0.077\scriptsize{(0.001)} &  0.085\scriptsize{(0.000)}   & 0.039\scriptsize{(0.002)}& 0.039\scriptsize{(0.032)} & \textbf{0.017\scriptsize{(0.024)}}    \\ 
\cmidrule{1-8}
\multirow{4}{*}{\emph{Random}} & \textit{Cap. Rate}$\uparrow$ & 0.277\scriptsize{(0.002)}  & 0.145\scriptsize{(0.019)}& 0.314\scriptsize{(0.018)}  &0.563\scriptsize{(0.016)} & 0.783\scriptsize{(0.272)} & \textbf{1.000\scriptsize{(0.000)}}     \\ 
\cmidrule{2-8}
& \textit{Cap. Step}$\downarrow$   & 639.0\scriptsize{(001.6)}   & 716.0\scriptsize{(012.0)}   & 620.0\scriptsize{(014.8)} &488.67\scriptsize{(008.7)} & 493.8\scriptsize{(183.6)} & \textbf{329.9\scriptsize{(060.6)}}    \\ 
\cmidrule{2-8}
& \textit{Coll. Rate}$\downarrow$   & 0.015\scriptsize{(0.001)}   & 0.018\scriptsize{(0.002)}   & 0.022\scriptsize{(0.002)} &0.020\scriptsize{(0.000)} & 0.168\scriptsize{(0.160)} & \textbf{0.011\scriptsize{(0.008)}}    \\ 
\cmidrule{1-8}
\multirow{4}{*}{\emph{Passage}} & \textit{Cap. Rate}$\uparrow$ & 0.733\scriptsize{(0.010)} &0.009\scriptsize{(0.003)} &0.229\scriptsize{(0.027)}  &0.818\scriptsize{(0.027)} & 0.617\scriptsize{(0.275)} &  \textbf{1.000\scriptsize{(0.000)}}     \\ 
\cmidrule{2-8}
& \textit{Cap. Step}$\downarrow$   & 552.7\scriptsize{(005.0)} & 799.3\scriptsize{(000.5)}   & 750.3\scriptsize{(007.4)} & 533.7\scriptsize{(008.8)} & 541.0\scriptsize{(144.9)} &\textbf{329.6\scriptsize{(076.9)}}    \\ 
\cmidrule{2-8}
& \textit{Coll. Rate}$\downarrow$   & \textbf{0.000\scriptsize{(0.000)}} & 0.001\scriptsize{(0.000)}   & 0.001\scriptsize{(0.000)} & {0.001\scriptsize{(0.000)}} & 0.013\scriptsize{(0.008)} &{0.006\scriptsize{(0.008)}}    \\ 
\bottomrule
\end{tabular}}}
\caption{Results of all methods in test scenarios. Our approach significantly outperforms all baselines in test scenarios.}
\vspace{-7pt}
\label{tab:main}
\end{table*}

\subsection{Experiment Setting}
We use OmniDrones \cite{xu2023omnidrones}, a high-speed UAV simulator, to construct the task. The environment (Fig.~\ref{fig:snapshot}) is a 0.9-radius, 1.2-height arena with 4–5 randomly placed cylindrical obstacles (radius 0.1, height 1.2). Collisions are avoided by maintaining a minimum distance of 0.07 between UAVs and obstacles. The setup includes 3 quadrotors (max speed 1.0) pursuing an evader (speed 1.3), with a capture radius of 0.3 and a maximum episode length of 800 steps.

\subsection{Evaluation}
\begin{figure}
\centering
\centering
\includegraphics[width=0.48\textwidth]{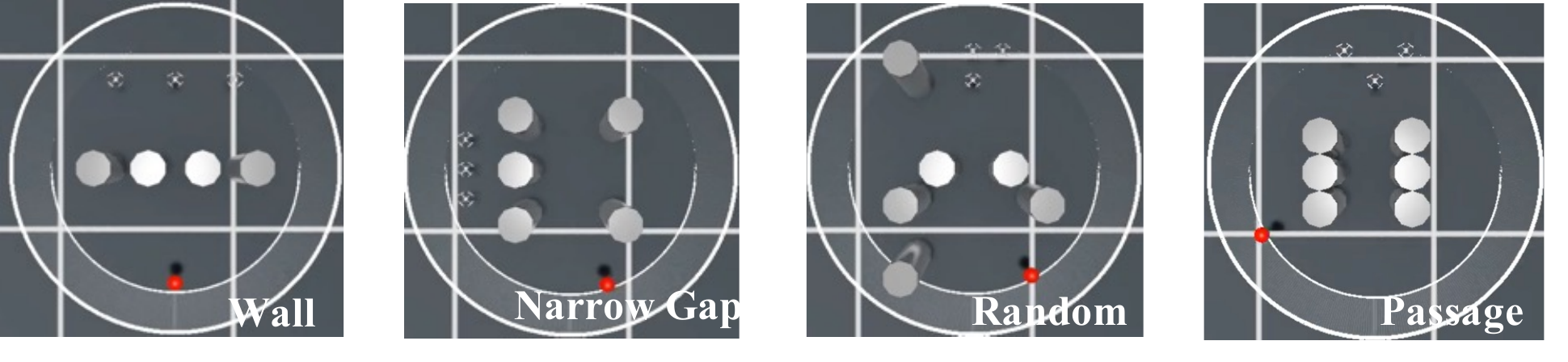}
\caption{Illustration of four test scenarios.}
\vspace{-7mm}
\label{fig:evaluation}
\end{figure}

\textbf{Evaluation Scenarios.} We design four test scenarios (Fig.~\ref{fig:evaluation}) to comprehensively evaluate our method. The within-distribution scenarios, \emph{Wall} and \emph{Narrow Gap}, challenge pursuit strategies: \emph{Wall} restricts access to lateral paths, while \emph{Narrow Gap} allows the evader to exploit a speed advantage along the perimeter. For out-of-distribution (OOD) evaluation, \emph{Random} tests autonomous exploration by hiding the evader initially, and \emph{Passage} requires strategic interception across three escape routes.

\textbf{Evaluation Metrics.} We use three statistical metrics to evaluate pursuit strategies: Capture Rate, Capture Step, and Collision Rate. Each experiment is averaged over 300 testing episodes (100 per seed). {All numerical values are reported as mean (standard deviation).}
\begin{itemize}
\item \textbf{Capture Rate: }Ratio of successful episodes where the evader is captured before the maximum episode length. Higher values indicate better performance.
\item \textbf{Capture Step: }Average timestep of first capture. If not captured, the maximum episode length is recorded. Lower values indicate quicker capture.
\item \textbf{Collision Rate: }Average number of collisions between quadrotors and obstacles or other quadrotors per episode length. Lower values indicate safer pursuit.
\end{itemize}

\subsection{Baselines}
We challenge our method with three heuristic methods (Angelani, Janosov, and APF) and two RL-based methods (DACOOP and MAPPO).

\begin{itemize}
    \item \textbf{Angelani}~\cite{angelani2012collective}: Pursuers are attracted to the nearest particles of the opposing group, i.e., the evader.
    \item \textbf{Janosov}~\cite{janosov2017group}: Janosov designs a greedy chasing strategy and collision avoidance system that accounts for inertia, time delay, and noise.
    \item \textbf{APF}~\cite{koren1991potential}: APF guides pursuers to a target position by combining attractive, repulsive, and inter-individual forces. 
    \item
    \textbf{DACOOP}~\cite{zhang2022multi}: DACOOP employs RL to adjust the primary hyperparameters of APF, enabling effective adaptation to diverse scenarios. For a fair comparison, we replace D3QN~\cite{wang2016dueling} in the original paper with a SOTA MARL algorithm, MAPPO.
    \item
    \textbf{MAPPO}~\cite{yu2022mappo}:  MAPPO is a SOTA multi-agent reinforcement learning algorithms for general cooperative problems. We regard MAPPO as a strong baseline for policy training, particularly in tasks where no additional specialized techniques are employed.
\end{itemize}

For heuristic methods, we perform a grid search on hyperparameters in the training scenarios. For DACOOP, we carefully tune the hyperparameters and use the best results. Baselines, treating the quadrotor as a point mass, outputs velocity commands executed by a velocity controller.

\begin{figure}[t]
\begin{minipage}{0.48\textwidth}
\centering
\subcaptionbox{Capture rate.\label{fig:radius_rate}}{\includegraphics[width=0.45\textwidth]{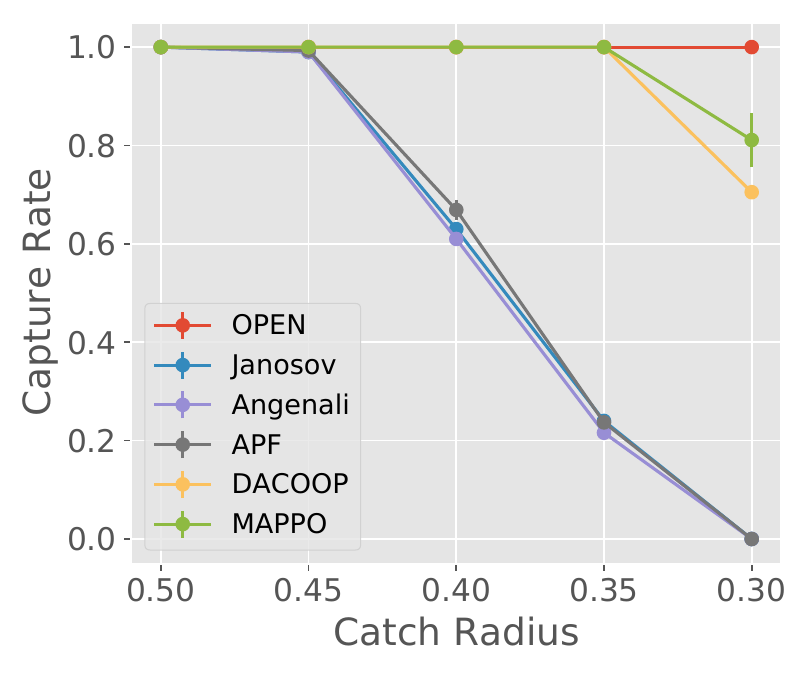}}
\subcaptionbox{Capture timestep.\label{fig:radius_timestep}}{\includegraphics[width=0.45\textwidth]{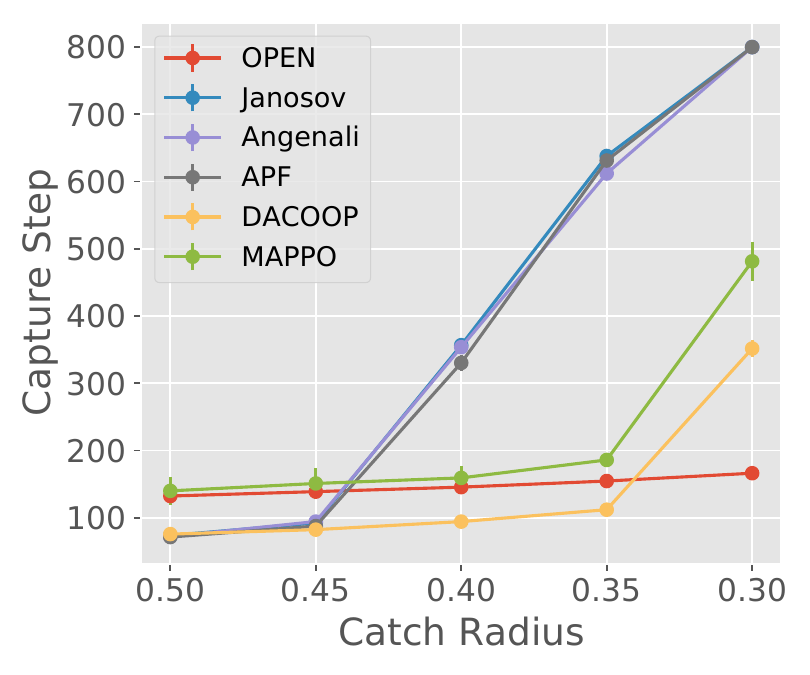}}
\caption{Results of all methods in a obstacle-free scenario. 
}\label{fig:empty}
\vspace{-4mm}
\end{minipage}
\end{figure}

\vspace{-3mm}
\subsection{Main Results in Simulation}
\textbf{Quantitative Results. } Tab.~\ref{tab:main} presents performance for the four test scenarios, focusing on corner cases. In \emph{Wall}, our method achieves over $98\%$ capture rate, with the lowest collision rate and fewest capture steps. In \emph{Narrow Gap}, it maintains a $100\%$ capture rate and the lowest collision rate, despite slightly higher capture steps than MAPPO. These within-distribution results highlight our method's effectiveness in corner cases and efficient agent cooperation. In \emph{Random} and \emph{Passage}, our method achieves $100\%$ capture rates, outperforming baselines ($78.3\%$ and $81.8\%$), and requires the fewest capture steps. While the collision rate in \emph{Passage} is slightly higher ($0.6\%$), the results demonstrate strong generalization to unseen scenarios. 

Given the poor baseline performance, we further test the algorithms in a simpler, obstacle-free scenario. As shown in Fig.~\ref{fig:empty}, baselines exhibit a sharp decline in capture rate as the capture radius decreases, underscoring the task's difficulty. A smaller radius demands more agile UAV behavior, requiring rapid formation adjustments to block all evader escape routes. However, heuristic algorithms and DACOOP, which rely on force-based adjustments for obstacle avoidance and pursuit, inherently limit UAV agility. When the capture radius is small (e.g., 0.3), MAPPO struggles to learn an optimal capture strategy, with the success rate converging to approximately 80\%. In contrast, our method maintains high capture rates with only a slight increase in capture steps as the radius shrinks, demonstrating the superior cooperative capture capability of pure RL strategies.

\begin{figure}
\centering
\centering
\includegraphics[width=0.48\textwidth]{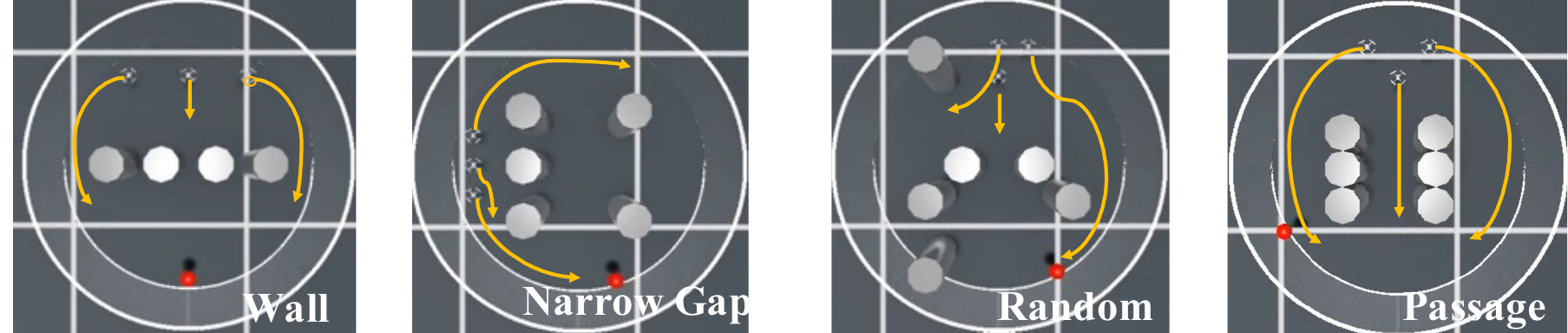}
\caption{Demonstration of the pursuit strategies.}
\vspace{-6mm}
\label{fig:behavior}
\end{figure}
\textbf{Behavior Analysis. }
As shown in Fig.~\ref{fig:behavior}, we identify four emergent behaviors in test scenarios, highlighting our strategy's cooperative pursuit capabilities (see supplementary video). In \emph{Wall}, our method employs a double-sided surround strategy: one quadrotor monitors while two flank the evader, unlike baselines hindered by frontal obstacles. In \emph{Narrow Gap}, our method intercepts the evader by taking shortcuts, unlike baselines that only follow. In \emph{Random}, guided by predicted trajectories, our method quickly locates the evader behind obstacles, while baselines fail initially. In \emph{Passage}, our method divides quadrotors to block all escape routes, whereas baselines greedily approach, leaving routes open.

\vspace{-3mm}
{\subsection{Ablation Studies}}
\label{sec:ab}

\begin{figure}[t]
\begin{minipage}{0.24\textwidth}
\centering
\subcaptionbox{Training curves.\label{fig:ab}}
{\includegraphics[width=\textwidth]{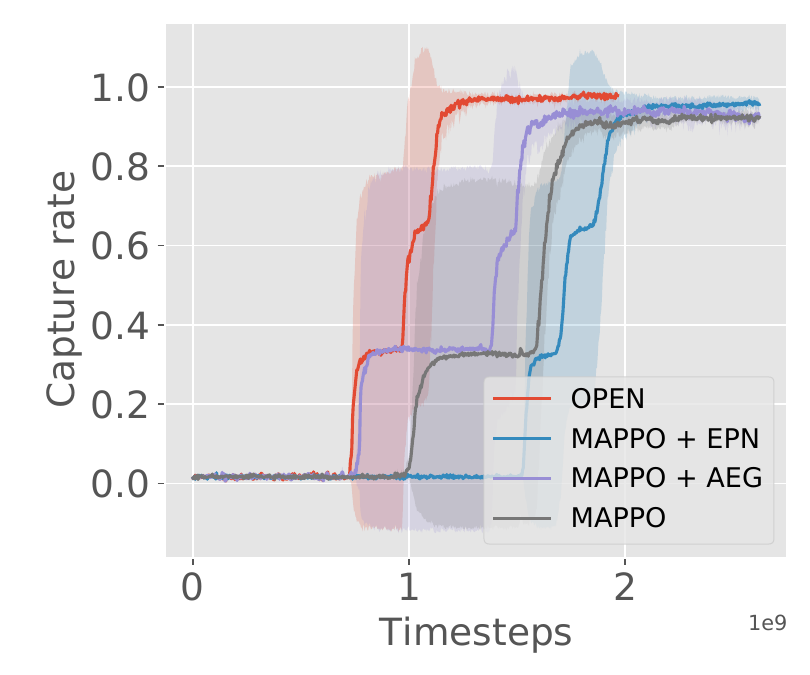}}
\end{minipage}
\begin{minipage}{0.24\textwidth}
\subcaptionbox{Policy Gen. across $v_e$. \label{fig:speed}}{\includegraphics[width=\textwidth]{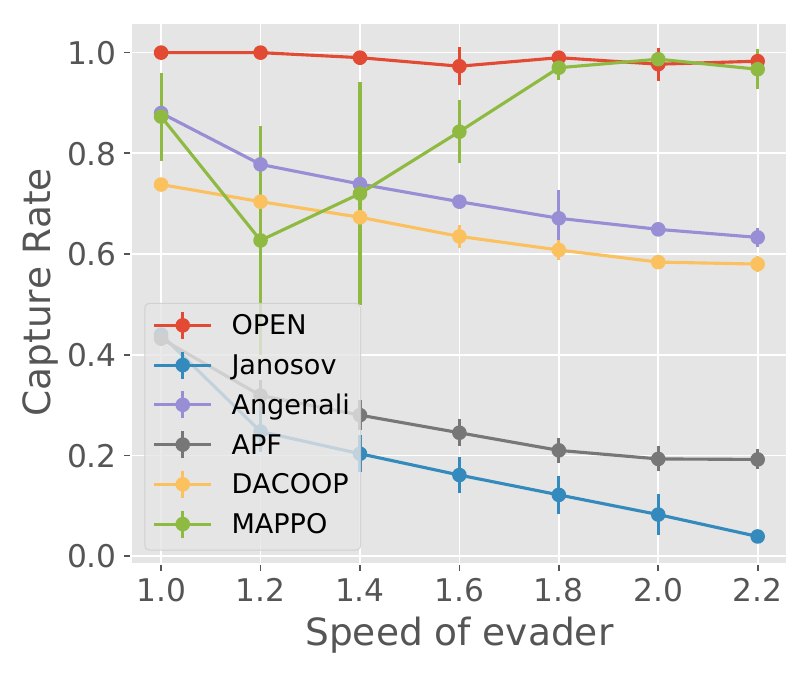}}
\end{minipage}
\caption{{Ablation studies. (a) Performance of {\name} and baselines on training tasks. (b) Policy generalization across evader speeds in the \emph{Passage} scenario.}}
\end{figure}

\begin{table}
\begin{minipage}{0.5\textwidth}
\centering
\footnotesize
\begin{tabular}{ccccc}
\toprule
Scenarios& \emph{Wall} & \emph{Narrow Gap} & \emph{Random} & \emph{Passage} \\ 
\midrule
Pred. Error&0.054\scriptsize{(0.013)}  & 0.084\scriptsize{(0.007)} & 0.109\scriptsize{(0.014)} & 0.081\scriptsize{(0.036)}  \\ 
\bottomrule
\end{tabular}
\caption{{Prediction errors of our method in four test scenarios.}}
\end{minipage}
\vspace{-5mm}
\end{table}

\begin{table}[t]
\begin{minipage}{0.5\textwidth}
    \centering
\footnotesize
{
{
\resizebox{0.95\textwidth}{!}{\begin{tabular}{cccccc}
\toprule
Methods& Metric & {\name} & MAPPO + EPN & MAPPO + AEG &MAPPO \\ 
\midrule
\multirow{4}{*}{\emph{Random}}& \textit{Cap. Rate}$\uparrow$ & \textbf{1.000\scriptsize{(0.000)}
}& \textbf{1.000\scriptsize{(0.000)}} & 0.967\scriptsize{(0.025)} & 0.783\scriptsize{(0.272)} \\ 
\cmidrule{2-6}
& \textit{Cap. Step}$\downarrow$& 329.9\scriptsize{(060.6)} & \textbf{301.0\scriptsize{(028.7)}} & 477.1\scriptsize{(044.8)} & 493.8\scriptsize{(183.6)}\\ 
\cmidrule{2-6}
& \textit{Coll. Rate}$\uparrow$ & \textbf{0.011\scriptsize{(0.008)}} & 0.012\scriptsize{(0.009)} & 0.107\scriptsize{(0.110)} & 0.168\scriptsize{(0.160)}\\ 
\cmidrule{1-6}
\multirow{4}{*}{\emph{Passage}}& \textit{Cap. Rate}$\uparrow$ & \textbf{1.000\scriptsize{(0.000)}} & 0.886\scriptsize{(0.120)} & 0.727\scriptsize{(0.193)} & 0.617\scriptsize{(0.275)}\\ 
\cmidrule{2-6}
& \textit{Cap. Step}$\downarrow$& \textbf{329.6\scriptsize{(076.9)}} & 427.6\scriptsize{(123.0)} & 602.2\scriptsize{(041.1)} & 541.0\scriptsize{(144.9)}\\ 
\cmidrule{2-6}
& \textit{Coll. Rate}$\uparrow$ & \textbf{0.006\scriptsize{(0.008)}} & 0.013\scriptsize{(0.010)} & 0.216\scriptsize{(0.292)} & 0.013\scriptsize{(0.008)}\\ 
\bottomrule
\end{tabular}}}}
\caption{{Comparison of {\name} and variants in OOD scenarios.}}\label{tab:ood}   
\end{minipage}
\end{table}

We conduct ablation studies on the core modules of our method. The baseline, denoted as ``MAPPO'', represents the naive implementation without the \emph{Evader Prediction-Enhanced Network} (EPN) and \emph{Adaptive Environment Generator} (AEG) modules. The variant ``MAPPO + EPN'' incorporates the \emph{Evader Prediction-Enhanced Network} module into the MAPPO framework, while ``MAPPO + AEG'' combines MAPPO with the \emph{Adaptive Environment Generator} module. 

\subsubsection{Performance on training tasks}
As shown in Fig.~\ref{fig:ab}, our method demonstrates the highest capture performance and improves sample efficiency by more than 50\% compared to MAPPO. ``MAPPO'' achieves only a 90\% capture rate with over 2.0 billion samples, highlighting the challenging nature of deriving an RL policy that considers UAV dynamics and performs well across diverse scenarios. And as shown in Tab.~\ref{tab:main}, MAPPO's average performance on the training tasks is suboptimal, leading to significantly worse performance in corner cases such as the \emph{Wall} and \emph{Narrow Gap} scenarios.

For ``MAPPO + EPN'', the training curves show that while the EPN module slightly increases training data requirements, it improves final performance over MAPPO. This is due to the additional computational overhead required to train the \emph{Evader Prediction-Enhanced Network}. Once the network converges and accurately predicts the evader's future position, ``MAPPO + EPN'' achieves superior performance. We also report the average prediction error (Pred. Error) of the evader's next position within an episode for the four test scenarios. Even in the \emph{Random} scenario, where the evader is initially invisible, the prediction error is comparable to the UAV size and much smaller than the capture radius, validating the predictions. For ``MAPPO + AEG'', the training curves show that the AEG module significantly improves training efficiency and slightly enhances final performance compared to MAPPO.

\subsubsection{Performance on out-of-distribution scenarios}
We further evaluate our method and its variants in out-of-distribution scenarios. As shown in Tab.~\ref{tab:ood}, our method achieves the highest capture rate and the lowest collision rate in the \emph{Random} scenario, with capture timesteps comparable to``MAPPO + EPN''. In the \emph{Passage} scenario, our method consistently outperforms all variants. The EPN module provides UAVs with the capability to predict the evader's strategy, allowing them to plan captures based on the evader's future trajectory. This significantly reduces the impact of OOD scenarios on policy performance. Additionally, the AEG module enhances the policy's ability to address corner cases, promoting better generalization across diverse environments. By integrating these two modules, our method surpasses all baseline variants.

\subsubsection{Policy generalization across evader speeds $v_e$}
As shown in Fig.~\ref{fig:speed}, we evaluate the sensitivity of {\name} and baselines to varying evader speeds in the \emph{Passage} scenario. Unlike obstacle-free settings, \emph{Passage} allows the evader to exploit obstacle occlusion and speed advantages. As speed increases, heuristic baselines and DACOOP show significant capture rate drops due to poor early interception. As the evader's speed increases, MAPPO's capture rate first drops then rises, as it learns a three-way interception strategy similar to {\name}. At higher speeds, the evader is more likely to be caught in its blind spot during turns, lacking time to adjust direction. In contrast, our method maintains robust performance across speeds, achieving high capture rates for faster evaders without retraining, despite being trained only for $v = 1.3$.

\vspace{-3mm}
\subsection{Real-World Experiments}
\begin{table}[t]
\centering
\footnotesize
{
\resizebox{0.45\textwidth}{!}{
{\begin{tabular}{cccccc}
\toprule
Methods & Envs. & \emph{Wall} & \emph{Narrow Gap}& \emph{Random} & \emph{Passage} \\ 
\midrule
\emph{w/o Smoothness} (\emph{oracle}) & Sim. & 258.6\scriptsize{(0.00)}  & 478.0\scriptsize{(0.00)} & 361.9\scriptsize{(0.00)}  & 362.0\scriptsize{(0.00)}   \\ 
\cmidrule{1-6}
\emph{w/o Smoothness} & Real. & fail  & fail & fail & fail   \\ 
\cmidrule{1-6}
\emph{One-stage} & Real. & / & /  & / & / \\
\cmidrule{1-6}
\emph{Two-stage} & Real. & 320.2\scriptsize{(26.8)}& 441.7\scriptsize{(82.1)}& 385.2\scriptsize{(29.4)}& 377.5\scriptsize{(49.2)}\\
\bottomrule
\end{tabular}}}}
\caption{Capture steps of all methods in four test scenarios. "/" notes that the \emph{One-stage} fails to generate effective pursuit strategies in the simulation and we do not conduct experiments in the real-world.}\label{tab:real}
\vspace{-5mm}
\end{table}
We deploy the policy on three real CrazyFlie 2.1 quadrotors, each with a maximum speed of 1.0 m/s. A motion capture system provides the quadrotors' states, while a virtual evader's true states are input into the actor and evader prediction network upon detection. The actor and evader prediction network run on a local computer, sending CTBR control commands at 100 Hz to the quadrotors via radio. {We align the initial conditions of the simulation with real-world to ensure fair comparison. All real experiments are repeated 5 times and recorded as the mean and standard deviation.}


{We compare the two-stage reward refinement (\emph{Two-stage}) with its variants. We define the first stage of \emph{Two-stage} as \emph{w/o Smoothness}, training only with task rewards. So the simulation results of \emph{w/o Smoothness} can be considered as \emph{oracle}. \emph{One-stage} denotes training with both smoothness and task rewards. As shown in Tab.~\ref{tab:real}, the results demonstrate that \emph{w/o Smoothness} fails completely in real-world experiments due to overly aggressive pursuit strategies which lead to unstable behaviors and UAV crashes. \emph{One-stage} causes severe exploration challenges in MARL training, resulting in no policy improvement during training and yielding only trivial stationary strategies. In contrast, only the \emph{Two-stage} achieves successful real-world deployment while preserving capture performance approximately consistent with simulation results.}


\section{CONCLUSION}\label{sec:conclusion}
We propose a feasible RL policy for online planning in unknown environments, enabling real-world deployment on Crazyflie quadrotors for multi-UAV pursuit-evasion. To promote generalization, we introduce an Adaptive Environment Generator for automatic curriculum generation and a Prediction-Enhanced Network for cooperative capture. Our method outperforms baselines in simulation and generalizes well to unseen scenarios. A two-stage reward refinement ensures physical feasibility and enables zero-shot real-world transfer.

Our method still has some limitations. The policy supports varying drone numbers but still needs retraining for performance. Moreover, future work will explore vision-based methods for unstructured environments.


\bibliographystyle{IEEEtran}
\bibliography{references}

\end{document}